\newif\ifanonymous \anonymousfalse
\newif\ifdraft \draftfalse
\newif\ifarxiv \arxivtrue
\newif\ifappendix \appendixtrue

\ifarxiv
    \ifanonymous
    \documentclass[acmtog,anonymous,authorversion,nonacm]{acmart_arxiv/acmart}
    \else
    \documentclass[acmtog,authorversion,nonacm]{acmart_arxiv/acmart}
    \fi

\else
    \documentclass[acmtog,anonymous,review]{acmart}
\fi

\acmSubmissionID{249}

\usepackage{booktabs} %

\usepackage[ruled]{algorithm2e} %

\SetAlFnt{\small}
\SetAlCapFnt{\small}
\SetAlCapNameFnt{\small}
\SetAlCapHSkip{0pt}

\acmJournal{TOG}

\usepackage{cleveref}
\usepackage{makecell}
\usepackage{soul}
\usepackage{diagbox}
\usepackage{subfigure}

\crefname{section}{Sec.}{Secs.}
\Crefname{section}{Section}{Sections}
\Crefname{table}{Table}{Tables}
\crefname{table}{Tab.}{Tabs.}
\Crefname{figure}{Figure}{Figures}
\crefname{figure}{Fig.}{Figs.}

\ifdraft
\newcommand{\todo}[1]{{\color{red} #1}}

\newcommand{\src}[1]{\textcolor{violet}{SR: #1}}
\newcommand{\sr}[1]{{\color{violet} #1}}

\newcommand{\gtc}[1]{\textcolor{cyan}{GT: #1}}

\newcommand{\gtdel}[1]{{\color{cyan} \st{#1}}}
\newcommand{\dcc}[1]{{\color{red}\textbf{DC:} #1}}

\newcommand{\nsc}[1]{{\color{blue}\textbf{NS:} #1}}

\newcommand{\igc}[1]{{\color{green}\textbf{IG:} #1}}

\newcommand{\rsc}[1]{{\color{orange}\textbf{RS:} #1}}

\newcommand{\abc}[1]{{\color{purple}\textbf{AB:} #1}}

\newcommand{\ofc}[1]{{\color{olive}\textbf{OF:} #1}}

\else
\newcommand{\todo}[1]{}
\newcommand{\src}[1]{}
\newcommand{\sr}[1]{#1}

\newcommand{\gtc}[1]{}

\newcommand{\gtdel}[1]{}

\newcommand{\dcc}[1]{}

\newcommand{\nsc}[1]{}

\newcommand{\igc}[1]{}

\newcommand{\rsc}[1]{}

\newcommand{\abc}[1]{}

\newcommand{\ofc}[1]{}

\fi

\newcommand{\algoname}{MoMo\xspace}  %
\newcommand{\algonamelong}{\textbf{Mo}nkey See, \textbf{Mo}nkey Do}
\newcommand{\benchname}{MTB\xspace}
\newcommand{\benchnamelong}{\textbf{M}otion \textbf{T}ransfer \textbf{B}enchmark\xspace}
\newcommand{\attnname}{mixed-attention\xspace}

\newcommand{\tracked}{leader\xspace}  %
\newcommand{\Tracked}{Leader\xspace}  %
\newcommand{\trackedidx}{\text{ldr\xspace}}
\newcommand{\follower}{follower\xspace} %
\newcommand{\Follower}{Follower\xspace}
\newcommand{\followeridx}{\text{flw\xspace}}
\newcommand{\outidx}{\text{out\xspace}}
\newcommand{\out}{output\xspace}  
\newcommand{\Out}{Output} 

\newcommand{\outline}{outline\xspace}
\newcommand{\motifs}{motifs\xspace}

\newcommand{\R}{\mathbb{R}} %
\newcommand{\E}{\mathbb{E}} %

\newcommand{\mll}{\mathcal{L}}

\newcommand{\nn}{\mathcal{N}}

\newcommand{\Loss}{\mll}

\newcommand{\abs}[1]{\lvert #1 \rvert}
\newcommand{\norm}[1]{\lVert#1\rVert}

\makeatletter
\ifundef{\onedot} 
{
\DeclareRobustCommand\onedot{\futurelet\@let@token\@onedot}
\def\@onedot{\ifx\@let@token.\else.\null\fi\xspace}
}
{}
\ifundef{\eg} {\def\eg{\emph{e.g}\onedot}} {} \ifundef{\Eg} {\def\Eg{\emph{E.g}\onedot}} {} 
\ifundef{\ie} {\def\ie{\emph{i.e}\onedot}} {} \ifundef{\Ie} {\def\Ie{\emph{I.e}\onedot}} {} 
\ifundef{\cf} {\def\cf{\emph{cf}\onedot}} {} \ifundef{\Cf} {\def\Cf{\emph{Cf}\onedot}} {} 
\ifundef{\etc} {\def\etc{\emph{etc}\onedot}} {} \ifundef{\vs} {\def\vs{\emph{vs}\onedot}} {} 
\ifundef{\wrt} {\def\wrt{w.r.t\onedot}} {} \ifundef{\dof} {\def\dof{d.o.f\onedot}} {} 
\ifundef{\iid} {\def\iid{i.i.d\onedot}} {}  \ifundef{\wolog} {\def\wolog{w.l.o.g\onedot}} {} 
\ifundef{\etal} {\def\etal{\emph{et al}\onedot}} {} 
\ifundef{\supp} {\def\supp{%
\ifappendix%
Appendix%
\else
sup. mat\onedot
\fi
}} {} 
\makeatother

\makeatletter
\newcommand\myparagraph{\@startsection{paragraph}{4}{\z@}%
  {-.1\baselineskip \@plus -2\p@ \@minus -.2\p@}%
  {-3.5\p@}%
  {\bf\ACM@NRadjust{\@parfont\@adddotafter}}}
\makeatother

\begin{document}
\title[Monkey See, Monkey Do]{Monkey See, Monkey Do: 
Harnessing Self-attention in Motion Diffusion 
for Zero-shot Motion Transfer 
}

\ifarxiv
\author{Sigal Raab$^1$,\hspace{1em} Inbar Gat$^1$,\hspace{1em} Nathan Sala$^1$, \hspace{1em} Guy Tevet$^1$,\hspace{1em} Rotem Shalev-Arkushin$^1$, \\ \hspace{1em} Ohad Fried$^2$,\hspace{1em} Amit H. Bermano$^1$,\hspace{1em}  Daniel Cohen-Or$^1$}
\affiliation{%
\institution{Tel Aviv University, Israel$^1$ \hspace{0.3em} Reichman University, Israel$^2$}
}
\affiliation{
\institution{sigal.raab@gmail.com}
}
\fi  %

\renewcommand\shortauthors{Raab, S. et al.}

\begin{abstract}

Given the remarkable results of motion synthesis with diffusion models, a natural question arises: how can we effectively leverage these models for motion editing?
Existing diffusion-based motion editing methods overlook the profound potential of the prior embedded within the weights of pre-trained models, which enables manipulating the latent feature space; 
hence, they primarily center on handling the motion space.
In this work, we explore the attention mechanism of pre-trained motion diffusion models.
We uncover the roles and interactions of attention elements in capturing and representing intricate human motion patterns, and carefully integrate these elements to transfer a \tracked motion to a \follower one while maintaining the nuanced characteristics of the \follower, resulting in zero-shot motion transfer. 
Manipulating features associated with selected motions allows us to confront a challenge observed in prior motion diffusion approaches, which use general directives (\eg, text, music) for editing, ultimately failing to convey subtle nuances effectively.
Our work is inspired by how a monkey closely imitates what it sees while maintaining its unique motion patterns; hence we call it \emph{\algonamelong}, and dub it \emph{\algoname}. 
Employing our technique enables accomplishing tasks such as synthesizing out-of-distribution motions, style transfer, 
and spatial editing. 
Furthermore, diffusion inversion is seldom employed for motions; as a result, editing efforts focus on generated motions, limiting the editability of real ones. \algoname harnesses motion inversion, extending its application to both real and generated motions.
Experimental results show the advantage of our approach over the current art. 
In particular, unlike methods tailored for specific applications through training, our approach is applied at inference time, requiring no training.
\ifarxiv
Our webpage, which includes links to videos and code, can be found at \url{https://monkeyseedocg.github.io}.
\fi

\end{abstract}

\maketitle

\section{Introduction} \label{sec:intro}
\begin{figure}
    \centering
    \includegraphics[width=\columnwidth]{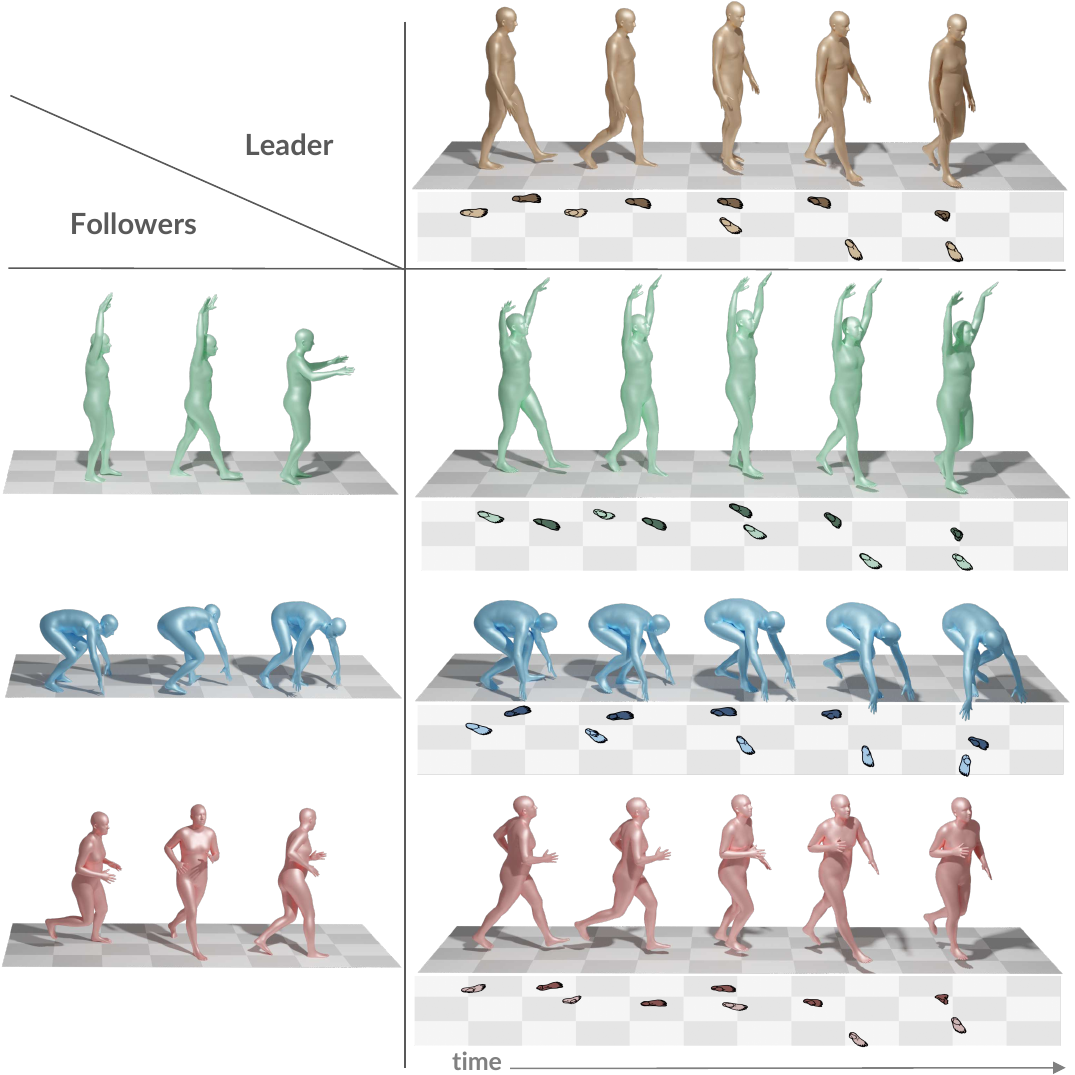}
    \vspace{-20pt}
    \caption{
    \textbf{Motion transfer.} The top row displays a \tracked performing a walking motion. The left column showcases sample frames of four {\follower}s, each engaged in a different motion. The central block presents the \out motion, where the \outline of the \tracked (\eg, leading leg) is transferred to the {\follower}s and integrated with their distinct motifs. Note the alignment of the steps for the \tracked and \out motions.  
    Our motion transfer is conducted by manipulating self-attention latent features in a zero-shot fashion.
    }
    \vspace{-12pt}
    \label{fig:teaser}
    \Description[]{}  %
\end{figure}

Human motion synthesis is a fundamental task, useful for various fields including robotics, autonomous driving, health care, gaming and animation.
Diffusion models~\cite{sohl2015deep,ho2020denoising} stand out as the prevailing synthesis paradigm across different modalities, such as imaging~\cite{saharia2022photorealistic}, video~\cite{ho2022video}, 3D point clouds~\cite{luo2021diffusion}, and also motion~\cite{tevet2023human,dabral2022mofusion}.
With the emergence of foundation models~\cite{bommasani2021opportunities},
it has been natural for some modalities, like imaging and video, to evolve towards zero-shot
editing~\cite{hertz2023prompt,geyer2024tokenflow}.
Such works typically depend on the prior information encoded in the weights of pre-trained models, which facilitate latent feature space manipulation. This capitalizes on a deep understanding of their intricacies, with attention layers playing a dominant role in these methods.

However, the prior information encoded in pre-trained \emph{motion} diffusion models remains largely unexplored.
Furthermore, a notable disparity exists between the imaging and motion domains. Images possess a regularized 2D spatial structure with a considerable number of degrees of freedom (DoF), whereas our motion 
is defined over a 3D human skeleton with a 1D temporal axis, and significantly fewer DoF. 
Therefore, insights regarding pre-trained imaging models do not directly apply to motion.

In the following, the term \emph{motion \outline} denotes a structured plan that organizes the key blocks involved in a specific motion (\eg, locomotion rhythm), and the term \emph{motion \motifs} denotes gestures or patterns (\eg, characteristic pose). These terms are further elaborated in \cref{sec:model}.

Human subjects exhibit highly expressive motions, containing a wealth of information. For example, a jumping motion can be combined with themes such as raising the arms or clapping the hands. A subtle change in each motion pattern would lead the viewer to a completely different impression. Similarly, discerning an individual's mood or age becomes possible by observing a short duration of their walking pattern. 
However, accurately conveying these subtle nuances through %
either high-level controls such as text or low-level controls such as joint trajectory
poses a challenge. Moreover, motion datasets are limited,
and acquiring real-world human motions representing various motifs solely through motion capture (MoCap) is expensive, slow, and unscalable. 

Diffusion works that facilitate motion editing~\cite{tevet2023human,goel2023iterative} modify motion features (\eg, rotation angles), hence are limited to fixed sets of joints or operations. Furthermore, they rely on textual control and thus cannot convey subtle motifs.
Works that \emph{do} manipulate latent features \cite{raab2023modi,tevet2022motionclip} precede the diffusion models era, lacking access to their rich priors.

In this work, we examine the \emph{self-attention} mechanism within the motion domain (\cref{sec:analysis}) and suggest an unpaired editing framework that 
transfers a \tracked motion to a \follower one while preserving the subtle \motifs of the \follower,
thereby reducing reliance on costly MoCap systems or overly general textual descriptions.
We name our work ``\algonamelong'' (dubbed \algoname), as it encapsulates the concept of directly transferring a motion \outline from a \tracked to a \follower while retaining the \follower's unique \motifs, much like a monkey mimicking observed behavior in a monkeyish manner.

\algoname offers a versatile motion transfer technique, facilitating tasks unified by the core concept of transferring motifs from one motion to another. Special cases include style transfer (\eg, transferring a walking motion to zombie walking), spatial editing (\eg, transforming a jumping motion into ``jumping with hands up''), action transfer (\eg, transitioning from walking to running), and out-of-distribution synthesis (\eg, generating a dancing gorilla). See \cref{fig:teaser,fig:qual_res} for more examples.
Our model operates at inference time without requiring optimization or training and can function seamlessly with any underlying motion synthesis backbone (i.e., foundation model) that utilizes self-attention, regardless of its specific architecture.

Our study hypothesizes that self-attention elements can capture complex motion patterns, 
delves into the distinct functionalities of attention elements and examines their interplay.
Inspired by research in the imaging domain \cite{hu2023animateanyone,cao2023masactrl, alaluf2023crossimagema}, we devise a method where the capability of a query $Q$ in a \tracked motion, is utilized to detect the most relevant keys $K$ from a \follower motion. 
Specifically, we calculate an attention score using the $Q$ from the \tracked motion and the $K$ from the \follower motion. 
This score is utilized to extract a weighted combination of values $V$ from the same \follower motion.
Consequently, a new motion is synthesized, incorporating the \outline from the \tracked motion, and the \motifs from the \follower motion, maintaining fidelity to both. 
In essence, the \tracked motion determines the \emph{``what''} and \emph{``when''}, and the \follower motion determines the \emph{``how''}. 
For example, an instance of the ``what'' could be ``a step forward with the right leg'', while an instance of the ``how'' can be ``running'', ``tiredly'', or ``with hands raised''.
\Cref{fig:corresp} illustrates the implicit semantic correspondence between the \tracked and \follower motions, which does not require additional supervision.

Our work stands as the sole approach capable of utilizing motion DDIM inversion within diffusion models
~\citep{dhariwal2021diffusion,song2020denoising}.
In the field of imaging, the integration of inversion with diffusion is widely used, facilitating the manipulation of real images~\cite{mokady2023null,huberman2023edit,garibi2024renoise}. 
However, in the motion domain, diffusion models typically avoid employing inversion. %
\algoname utilizes inversion, thereby enabling editing of both real and generated motions.

We introduce a comprehensive benchmark to evaluate our work. Our benchmark, named \benchname, will be made publicly available. It comprises selected motion pairs from the HumanML3D~\cite{guo2022generating} test set and is described in \cref{sec:benchmark}. Using \benchname, we compare our model to current state-of-the-art methods and demonstrate that none offer the same breadth of functionality as \algoname, which consistently outperforms them.

\section{Related Work} \label{sec:rel_work}

\subsection{Motion Synthesis}

\myparagraph{Multimodal synthesis}
\citet{petrovich2021actor,petrovich2022temos} incorporate a Transformer~\citep{vaswani2017attention} architecture for the tasks of action-to-motion and text-to-motion. T2M~\citep{guo2022generating}, T2M-GPT~\citep{zhang2023generating} and MotionGPT~\citep{jiang2024motiongpt} use VQ-VAE~\citep{van2017neural} to quantize motion, then sequentially synthesize it in the quantized space conditioned on text.
More recently, MDM~\citep{tevet2023human} and Mofusion~\citep{dabral2022mofusion} adapted the denoising diffusion framework~\citep{ho2020denoising} for motion synthesis and showed its merits for multimodal tasks such as action-to-motion, text-to-motion, and music-to-motion~\citep{tseng2023edge}. The MAS~\cite{kapon2023mas} algorithm extended diffusion to out-of-domain motions by leveraging video data.
\algoname excels by using a \emph{\follower} reference instead of an overly generalized text. 

\myparagraph{Spatial control and editing}
Motion Graphs~\citep{kovar2002motion,arikan2002interactive,lee2002interactive} are a popular data structure for traversing the pose space of a given motion dataset to synthesize new motion variations, often according to an input trajectory.
GANimator~\citep{li2022ganimator} and SinMDM~\citep{raab2024single} show that, analogously to images~\citep{shaham2019singan,nikankin2022sinfusion}, overfitting 
a single motion 
allows learning its internal distribution and synthesizing new variations of it with spatial and temporal variations.
Using diffusion, MDM enables both joints and temporal editing by adapting image diffusion inpainting~\citep{saharia2022palette,song2020score}.
Followup works demonstrate the merits of propagating control signals through the gradual denoising process for various applications such as long motion synthesis~\citep{zhange2023diffcollage,petrovich24stmc}, motion in-betweening~\citep{cohan2024flexible,xie2023omnicontrol}, and single joint control~\citep{shafir2024human,karunratanakul2023gmd}. 
ComMDM~\citep{shafir2024human} uses diffusion to synthesize two actors.   
InterGen~\citep{liang2024intergen} follows it and shows the effectiveness of cross-attention in the training process.
SINC~\citep{athanasiou2023sinc} and FineMoGen~\citep{zhang2024finemogen} use Large Language Models (LLMs) to break the text prompt to instruct each body part separately. 
\citet{goel2023iterative} use the coding skills of LLMs with predefined pose modifiers for frame editing, then blend them using diffusion.
Our approach enables spatial editing of the \emph{\tracked} motion as a special case of motion transfer.

\myparagraph{Style transfer}
One of the special cases enabled by \algoname is style transfer.
\citet{holden2016deep,holden2017fast} have suggested learning the motion manifold using an auto-encoder neural network. The latent space of the auto-encoder exposes semantic features of the motion, which enables motion stylization using the Gram matrix heuristic as suggested by \citet{gatys2015neural}. 
\citet{aberman2020unpaired} use the AdaIN heuristic to disentangle content and style as presented in StyleGAN~\citep{karras2019style}, followed by \citet{guo2024generative} and \citet{kim2024most}. 
Unlike \algoname, these models are trained on predefined styles and struggle to generalize.

\subsection{Attention Control in the Imaging Domain}

The latent information encapsulated in the attention layers of the popular UNet~\citep{ronneberger2015u} architecture is extensively used in the image domain to guide and control the denoising diffusion process.
PnP~\citep{pnpDiffusion2022}, MasaCtrl~\citep{cao2023masactrl} and CIA~\citep{alaluf2023crossimagema} show that the self-attention layers encode structural information that can be used to edit an image without losing its original composition.
Prompt-to-Prompt~\citep{hertz2023prompt} and Attend-and-Excite~\cite{chefer2023attend} show that certain aspects of the image can be edited by manipulating the cross-attention with the input text, without affecting the rest. 
\citet{patashnik2023localizing} and \citet{dahary2024yourself} are manipulating the self- and cross-attention layers to control the layout of the image and avoid semantic leakage between its different parts.
Tune-A-Video~\citep{wu2023tune}, TokenFlow~\citep{geyer2024tokenflow} and Q-NeRF~\citep{patashnik2024consolidating} observe that the attention query, $Q$, encodes the structure while the key and value, $K$ and $V$, encode the appearance, and use it for mutual editing of images preserving temporal and structural consistencies. Our work follows the latter, leveraging self-attention layers for motion editing. Unlike imaging works, our work uses layers from a transformer and not from a UNet.

\section{Model} \label{sec:model}

\begin{figure}
    \centering
    
    \includegraphics[width=\columnwidth]{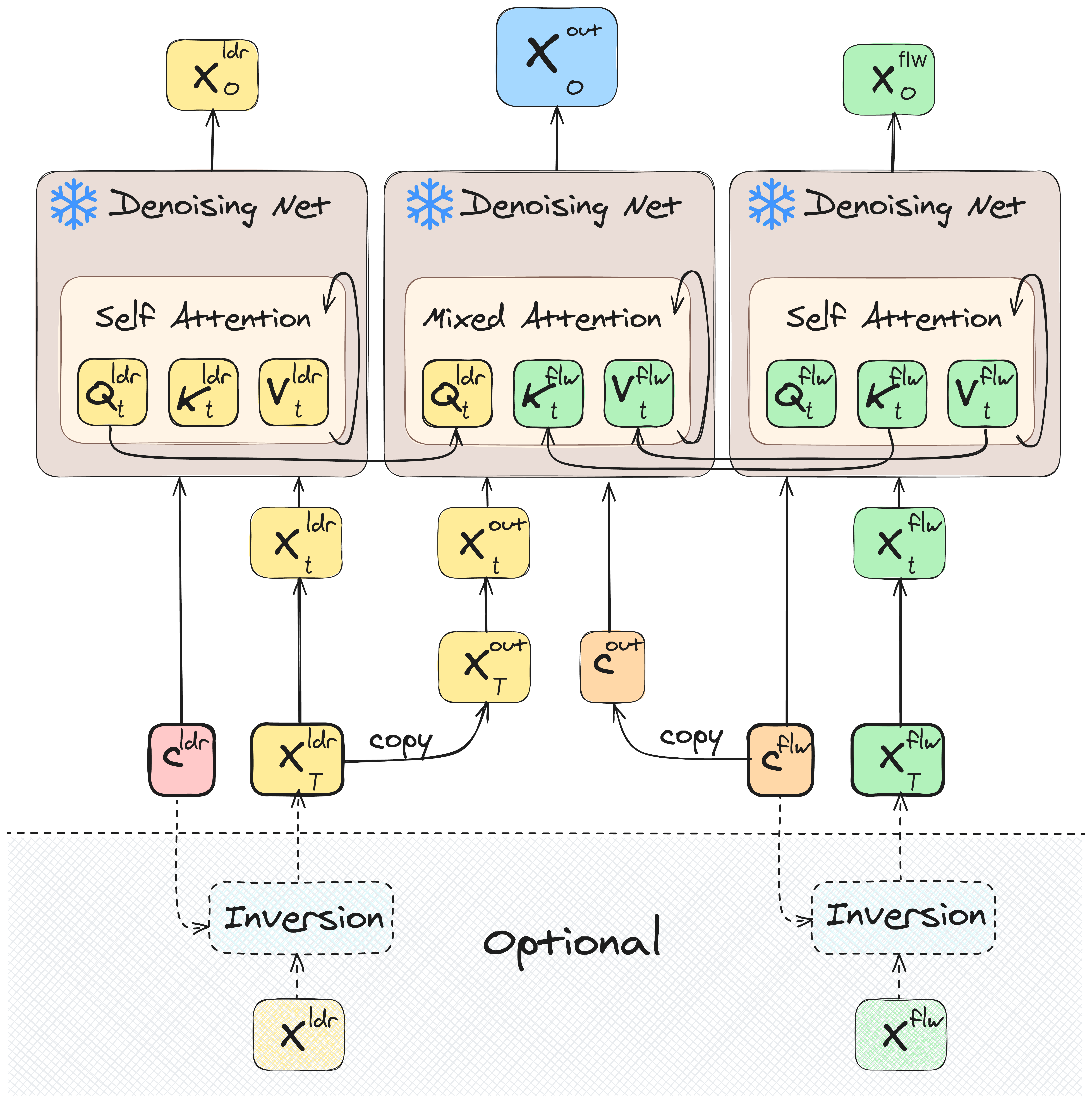}
    
    \caption{
    \textbf{The \algoname Pipeline.}
    The input to our model is two noisy tensors, $X_T^{\trackedidx}$ and $X_T^{\followeridx}$, produced by either inverting real motions or sampling a Gaussian noise.
    The two tensors represent \tracked and \follower motions, and are given along with their associated text prompts.
    We initialize our \out motion, $X_T^{\outidx}$, using the initial noise from the \tracked motion and pair it with the text prompt from the \follower motion.
    The three noised motions $X_t^{\trackedidx}$, $X_t^{\followeridx}$ and $X_t^{\outidx}$, are passed to the frozen denoising network at each timestep $t$, along with their prompts and with $t$.
    Within the denoising network, $X_t^{\outidx}$ undergoes \attnname by combining the query from the \tracked motion with the key and value from the \follower motion. Meanwhile, $X_t^{\trackedidx}$ and $X_t^{\followeridx}$ follow a standard diffusion process.
    }
    \label{fig:arch}
    \Description[]{}  %
\end{figure}

\begin{figure*}
    \centering
    \includegraphics[width=\linewidth]{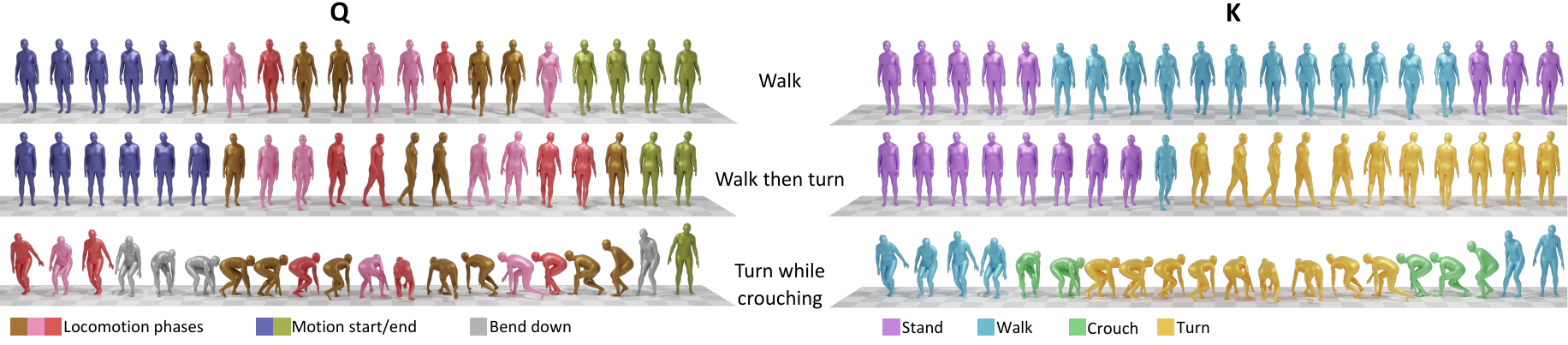}

    \vspace{-5pt}
    
    \caption{\textbf{Dominant features in Q vs. K.} 
    Each row depicts two copies of the same motion, showcasing the K-Means clustering of its $Q$ and $K$ features in the left and right columns, respectively. 
    Note how the features in $Q$ are dominated by the \outline, while those in $K$ are dominated by the motion \motifs. 
    In the $Q$ column, periodic steps share clusters, ignoring unique patterns. In the $K$ column, clusters are related to motion \motifs; thus, walking, turning while walking, and crouching while walking have distinct clusters.
    Temporal information is evident in the clusters of $Q$ but not in those of $K$. In the $Q$ column, the beginnings of the first two motions and the end of all three are highlighted by the colors of low and high frame numbers, respectively.
    }
    \vspace{-5pt}

    \label{fig:QK_analysis}
    \Description[]{}  %
\end{figure*}

This section suggests an editing framework that transfers the \outline of a \tracked motion to a \follower one while preserving the motion \motifs of the \follower.
Our unpaired framework operates at zero-shot, without requiring optimization or model training.

A motion \emph{outline} denotes a structured plan that arranges the essential movements necessary for a specific motion. It provides a visual blueprint for comprehending the sequence of actions and transitions needed to execute the motion effectively.
An example of a motion outline would be ``stand still on frames 10-20, step with right leg on frames 21-25'', etc. Note that the \outline is as general as possible; for example, the type of step (walk, run, hop) belongs to the motifs.
Motion \emph{motifs} include subtle nuances, gestures, or patterns that convey meaning and emotion. These motifs may repeat and vary, forming 
expressive motion sequences, and aiding in establishing visual themes and narratives. 
Consider various running motions, each with distinct motifs. These motions convey personal styles expressed by body angle, foot positioning, hand gestures, airborne duration, etc. Even with extensive prompt engineering, capturing every subtle motif remains unattainable. Conversely, incorporating motifs from a given motion ensures complete fidelity to that motion.

Our framework enables transitions across motions of different temporal lengths. The outlines of a leader motion can be transferred to multiple followers, as shown in Fig. 1, and multiple outlines can be separately applied to a single follower.

\subsection{Preliminaries} \label{sec:prelim}

\myparagraph{Motion Representation}
Let $N$ denote the number of frames in a motion sequence, and $F$ denote the length of the features describing a single frame, also known as pose. Finally, let $J$ denote the number of skeletal joints.
We adhere to the representation used in the HumanML3D dataset~\cite{guo2022generating}, where the features from all the joints are concatenated into a single large feature,
resulting in a motion representation $X\in \R^{N\times F}$. 
Details regarding the internal representation of the features can be found in the \supp.

\myparagraph{Self Attention}
We recap self-attention ~\citep{vaswani2017attention}, as it plays a key role in our framework. 
Let ${IH}$ (``input hidden'') be a latent tensor of features fed as input to a self-attention layer, and let $\hat{H}$ be the output of this layer. 
The elements query, key, and value, are calculated respectively by 
\begin{equation}
Q={IH}\cdot W_q^T+b_q,\quad K={IH}\cdot W_k^T+b_k,\quad  V={IH}\cdot W_v^T+b_v,
\end{equation}
where $(W_q,b_q)$, $(W_k,b_k)$, and $(W_v,b_v)$ are learned linear projections.

For each query vector $q_n\in Q$ at temporal location (\ie, frame) $n$, an attention score is computed based on all keys in $K$.
This score indicates the relevance of each key to the query $q_n$, assessing their similarity.
The attention scores are normalized through a softmax operation, which determines the weighting of each value in $V$, to be used for updating the features at location 
$n$. The weighted values are then aggregated to produce the output at each query location, via
\begin{equation}
    A_n  = \text{softmax} \left ( \frac{q_{n}\cdot K^T}{\sqrt{\abs{IH_n}}} \right ),\quad \quad \hat{h}_n  = A_n\cdot V,
\end{equation}
where $A_n$ is the normalized attention score at frame $n$, $\hat{h}_n\in \hat{H}$ is the self-attention result at frame $n$, and $\abs{IH_n}$ is the features number in frame $n$.
Finally, $\hat{H}$ is used as a residual and added to the input ${IH}$,
\begin{equation}
    {OH}  = {IH} + \hat{H},
\end{equation}
where ${OH}$ (``output hidden'') is the tensor passed to the next
layer.
We use multi-head attention and omit its notations for brevity.

\myparagraph{MDM, DDPM and DDIM}

Motion Diffusion Model (MDM)~\cite{tevet2023human} is a widespread model for human motion synthesis and editing. In our work we use a variation of it, hence we recap it here. MDM uses Denoising Diffusion Probabilistic Models (DDPM)~\cite{ho2020denoising}, which are trained to transform unstructured noise into samples from a specified distribution. This is achieved through an iterative process involving the gradual removal of small amounts of Gaussian noise.
Unlike MDM, in this work, we employ DDIM~\cite{song2021denoising} for inversion and deterministic inference, aiming to reconstruct inverted motions precisely to their original form.
A recap regarding DDPM and DDIM can be found in the \supp.

\subsection{Pipeline} \label{sec:pipeline}
\algoname employs a pre-trained and fixed motion diffusion model to
synthesize the \out motion $X^{\outidx}$ by applying the motion \outline of $X^{\trackedidx}$ onto $X^{\followeridx}$, where $X^{\trackedidx}$ and $X^{\followeridx}$ are either given (real) motions or generated ones (See \cref{fig:arch}). 

The input to our framework is two input noises, $X_T^{\trackedidx}$ and $X_T^{\followeridx}$ and their corresponding text prompts $c^{\trackedidx}$ and $c^{\followeridx}$, respectively. The input noises are either inverted from real motions using DDIM~\cite{song2021denoising} or sampled from a Gaussian distribution, in which case \algoname runs on generated motions.
The text prompts assist in controlling synthesis from noise and inverting the motions if inversion is used.
At each timestep $t$, we pass the three noised motions $X_t^{\trackedidx}$, $X_t^{\followeridx}$ and $X_t^{\outidx}$ to the denoising network.
$X_t^{\trackedidx}$ and $X_t^{\followeridx}$ go through a standard DDIM denoising, while 
$X_t^{\outidx}$ is denoised using our mixed-attention block,
 described next. 
Finally, \algoname 
produces the \out motion $X_{0}^{\outidx}$.

The denoising network in our pipeline can be any motion-denoising model that utilizes self-attention layers. In our experiments, we employ a variant of MDM as a backbone.
Note that, while related works in the imaging domain utilize the self-attention layers of a UNet~\cite{ronneberger2015u}, we leverage the self-attention layers of a Transformer~\cite{vaswani2017attention} due to the usage of MDM.

\subsection{Leveraging Self-attention}  \label{sec:mixed_attn}
Our proposed framework integrates self-attention components from both the \tracked and \follower motions into a single \out motion. 
In the imaging domain,
\citet{cao2023masactrl} have studied the self-attention layers of text-to-image denoising networks. They demonstrate that keeping the keys and values of these layers helps preserve the visual characteristics of objects when performing non-rigid manipulations on a given image. \citet{alaluf2023crossimagema} made further progress by combining structure and appearance from two images.

Inspired by these insights, this work illustrates the crucial functions of queries, keys, and values in encoding semantic motion information. We find that leveraging the queries, keys, and values from self-attention layers enables the transfer of semantic information across different motions.

In \cref{sec:analysis} we show that this approach enables the implicit transfer of motion patterns between semantically similar frames. 
More precisely, at each denoising step $t$, we use our mixed-attention block to inject the queries from the \tracked motion $X^\trackedidx$, and the keys and values from the \follower motion $X^{\followeridx}$ to the self-attention block of the \out motion $X^{\outidx}$, via 
\begin{equation}~\label{eq:mixed_attention}
    {OH}^{\outidx}  = {IH}^{\outidx} + \text{softmax} \left ( \frac{Q^{\trackedidx}\cdot K^{\followeridx^T}}{\sqrt{\abs{IH_n}}} \right ) V^{\followeridx}.
\end{equation}

\section{Understanding Self-attention Features} \label{sec:analysis}

In this section, we explore some of the prior information encoded in pre-trained motion diffusion models and identify useful attributes within it.
In the following, we demonstrate that (i) the \emph{queries} $Q$ establish a focal point for contextual determination, (ii) the \emph{keys} $K$ serve as a learned frame descriptor, enabling the model to assess the importance of different frames in the motion relative to a specific query, and (iii), the \emph{values} $V$ denote the contextual representation we seek to generate, guiding the model in shaping the features at each query's temporal location.
In particular, we show that while information regarding the \outline and the \motifs of the motion is contained in both $Q$ and $K$, the \outline information is more dominant in $Q$ and the \motifs information is more dominant in $K$. This key insight is presented in \cref{sec:QK_analysis} and serves as the foundation on which our mixed-attention block was built; to the best of our knowledge, this insight has not been explored in the imaging or motion domains.

\subsection{Distinguishing Between \texorpdfstring{$Q$ and $K$}{}} \label{sec:QK_analysis}

To understand the different roles of queries and keys, we extract their features from a chosen self-attention layer $\ell$ at diffusion step $t$, reduce the dimension to $d$ output channels by applying PCA, and group the frames into $m$ clusters using K-Means.
While $Q$ and $K$ contain both \outline and \motifs information, applying K-Means emphasizes the more dominant features.
\Cref{fig:QK_analysis} visualizes each cluster using a different color. Clustering $K$ shows that it is dominated by the motion \motifs and that similar \motifs are grouped into the same clusters. Clustering $Q$ depicts that it is dominated by the motion \outline. In particular, (i) periodic sub-motions, like steps, are grouped into the same cluster, dominating over motion \motifs, and (ii) the temporal signal is dominant.
From (i) we conclude that the $Qs$ in \emph{different} periodic motions share similar features, thus they can be ``understood'' by $Ks$ from other motions. This finding explains why our motion transfer works well.
Interestingly, $Q$ encodes the locomotion phases without explicitly learning it as in PFNNs~\cite{holden2017phase,starke2022deepphase}.
Our PCA is applied on $256$ generated motions with $128$ text prompts. We use $\ell\mbox{=}11$, $t\mbox{=}30$, $d\mbox{=}10$, and $m\mbox{=}10$.

\subsection{Correspondence via attention}  \label{sec:latent_nearest_neighbors}
\begin{figure}
    \centering
    \includegraphics[width=\columnwidth]{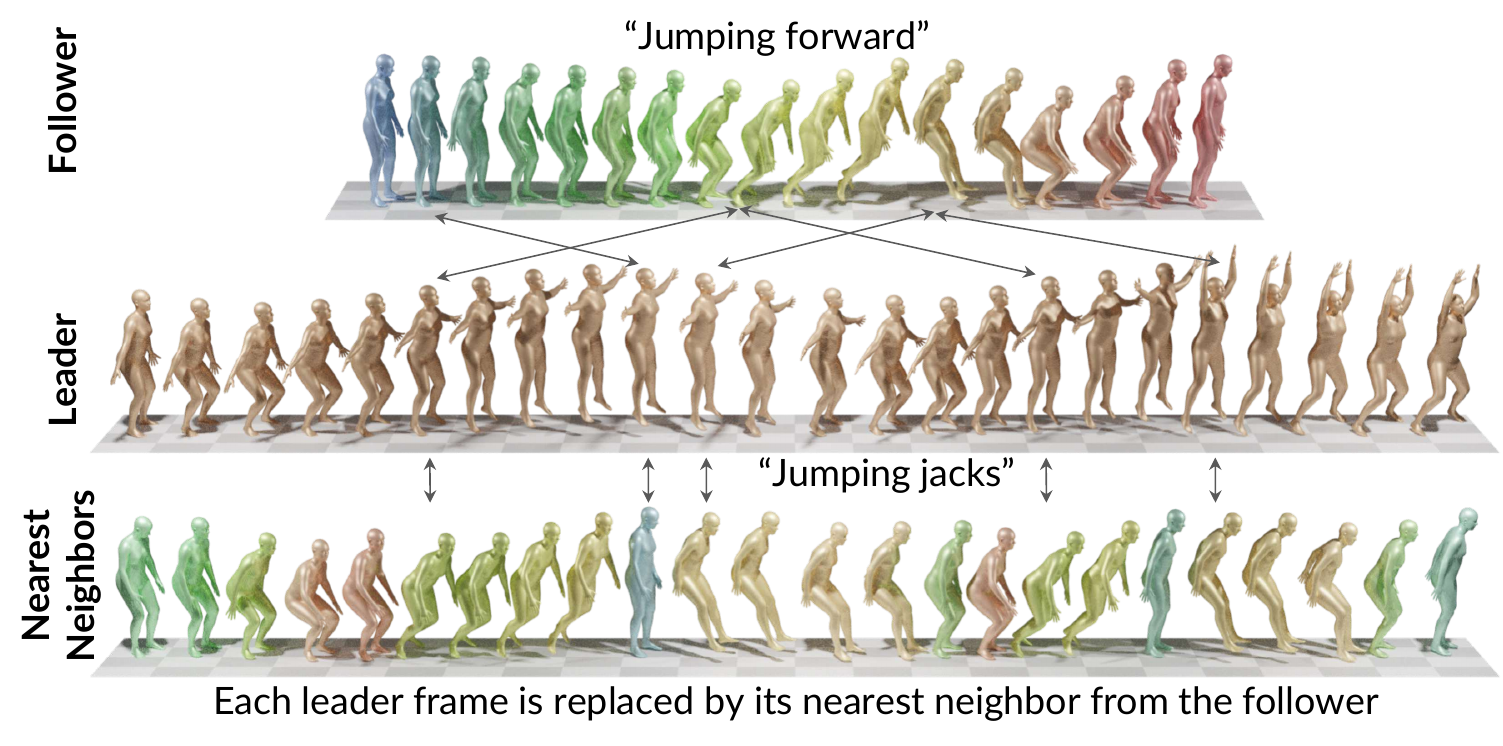}

    \vspace{-10pt}
    
    \caption{
    \textbf{Correspondence via attention.} 
    \Follower frames are color-coded according to consecutive indices (top row). 
    Nearest neighbor \follower frames (bottom) are the ones that achieve the highest \attnname ($Q^{\trackedidx}\cdot K^{\followeridx^T}$) activation, 
    shown respectively to \tracked's frames (middle row).
    These correspondences are semantically aligned, \eg, moving ``up'' and ``down'' sub-motions are consistently assigned with \follower moving ``up'' and ``down'' frames. 
    Some of the nearest neighbors are highlighted with arrows.
    }
    \vspace{-5pt}
    \label{fig:nn}
    \Description[]{}  %
\end{figure}

Given two entirely different motions, we show that the nearest neighbor from a \follower motion, for a query $Q$ from a \tracked motion, would be the frame that resembles it most in its \outline.
In \cref{fig:nn}, for every frame $n$
within the \tracked motion, we discern the frame in the \follower motion that elicits the greatest activation in the attention map $q_n^{\trackedidx}\cdot K^{\followeridx^T}$.
These pairings portray the transition from the \tracked to the \follower motion, highlighting the most significant activations. As depicted, these pairings exemplify alignment in the \outline, such as the synchronization of body movement (``up''/``down'') across frames. 
The attention maps in this section are computed for layer $10$ and diffusion step $50$.

\subsection{Attention Maps}  \label{sec:corresp_per_frame}
\begin{figure*}
    \centering
    \includegraphics[width=\linewidth]{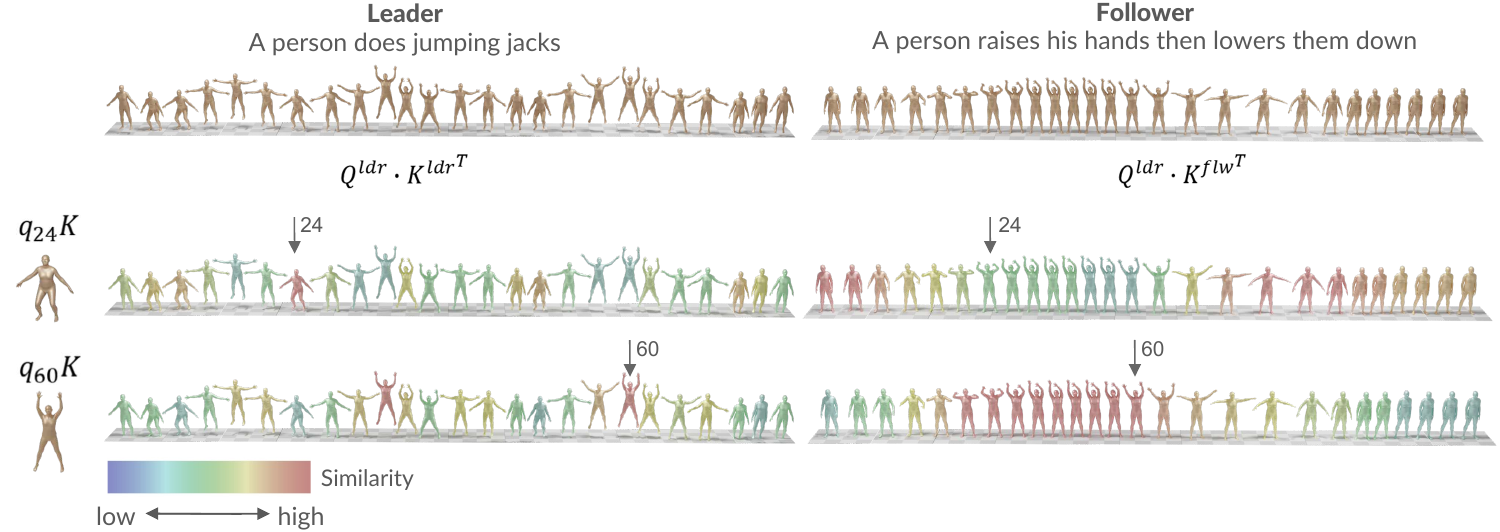}
    \vspace{-10pt}
    \caption{
    \textbf{Attention map per query.}
    In the left column, we display three copies of the \tracked; in the right column, we show copies of the \follower. The top copies depict the motions as they are, while the ones below highlight attention scores.
    We define two queries corresponding to different semantic temporal regions in the \tracked motion. Each query corresponds to a different pose, with varied arm direction or body stretch.
    Each motion column displays attention maps from a single layer, computed in different ways.
    In the left column, we present self-attention maps derived from queries and keys from the \tracked motion, causing each query to concentrate on semantically similar regions within that motion.
    The frame number related to each query is indicated with an arrow. 
    For example, the query in frame 24 focuses on a pose of ``standing low in an A pose'', in the \tracked motion. However, frame number 24 corresponds to an entirely different pose in the \follower motion in the right column.
    In the right column, we apply \algoname, aligning \tracked queries $Q^{\trackedidx}$ with \follower keys $K^{\followeridx}$. This way we ensure that each query from the \tracked motion aligns with semantically similar regions of the \follower motion.
    For instance, in frame 60, the query highlights the region where the character raises their arms. The frames with higher correspondence (red) in the right column also belong to characters raising their arms.
    }
    \vspace{-5pt}
    \label{fig:corresp}
    \Description[]{}  %
\end{figure*}

After computing attention maps in the previous section, which revealed the highest activation for each frame, we now delve into the complete attention maps for two selected frames.
Within each row in \cref{fig:corresp}, we focus on a temporal region identified by the number of its center frame.
In each column, we present the attention maps for each query location using various combinations of keys and queries. 
Naturally, when keys and queries from the same motion are multiplied ($Q^{\trackedidx}\cdot K^{\trackedidx^T}$), they produce high similarity scores for regions resembling the queried pose, especially the queried frame itself.
In contrast, in the \follower motion, the pose at the same frame number as the query may not be similar, making it unhelpful for retargeting the \tracked motion.
In the right column, we use \algoname to compute $Q^{\trackedidx}\cdot K^{\followeridx^T}$. 
As illustrated, the queries focus on semantically similar regions in the \follower motion. %

Note that these correspondences are made despite the two motions having different sequences of poses.
As a result, multiplying the attention maps with the follower values $V^{\followeridx}$, enables the transfer of the \outline of the \tracked motion onto the \follower's \motifs.

Finally, \cref{fig:attention_full_motion} depicts attention maps for the full motion length, providing visual insight into \tracked-\follower correspondence.

\section{Experiments} \label{sec:exp}

The results in this work are computed using the transformer decoder version of MDM. The exact hyperparameter values are detailed in the \supp.
In practice, \cref{eq:mixed_attention} is applied in diffusion steps 90 to 10 (out of 100) and in layers 2 to 12 (out of 12).

\subsection{Benchmark} \label{sec:benchmark}
To evaluate \algoname, we introduce
\emph{\benchnamelong}, dubbed \emph{\benchname}, of \tracked and \follower motion pairs, which will be made available.
\benchname is a subset of the HumanML3D~\cite{guo2022generating} test set and is straightforwardly filtered according to the textual prompts attached to the motions.
For the \tracked motions, we include motions that contain locomotion verbs such as ``run'' or ``walk''.
For the \follower motions, we include motions with text that contains words suggesting motion \motifs, such as ``chicken'' or ``clap''.
The word choices, shared in our \supp, are made straightforwardly and do not involve large language models.

We create pairs of \tracked and \follower motions by combining sentences from the respective groups in a Cartesian product. However, such a combination induces approximately $46K$ pairs, which is more than we need. 
To decrease the number of pairs, instead of using all the cross combinations of \tracked and \follower, we allow up to $20$ repetitions of each \follower sample and no repetitions for the \tracked samples.
In practice, we use $4$ \tracked search words, resulting in $842$ motions, and $17$ \follower search words, resulting in $55$ motions (some search terms result in less than three sentences). Altogether, \benchname includes $842$
(\tracked, \follower) motion pairs.

\subsection{Metrics} \label{sec:metrics}
We aim to assess the results based on three criteria: (a) the quality of the \out motions, (b) how closely the \out motions follow the \outline of the \tracked motion, and (c) how well the \out motions align with the \motifs of their associated \follower motions. 
To evaluate these aspects, we have chosen specific metrics. Criterion (a) is evaluated using FID, (b) is assessed by foot contact similarity, and criterion (c) is evaluated by R-precision and similarity to \follower locations and rotations. Details about these metrics are provided in the \supp.

\subsection{Baselines} \label{sec:baselines}
No existing baseline encompasses all aspects of motion transfer addressed by \algoname. 
Hence, we compare \algoname with several works, each representing a different special case. 
We demonstrate that \algoname offers superior generalization capabilities compared to these works. 
Additionally, most prior art is tailored for specific applications through training, and often requires paired data. In contrast, \algoname is unpaired and accomplishes its functionalities during inference without necessitating specific training or optimization.

\myparagraph{Motion transfer Using Na\"ive Nearest Neighbor}
To illustrate the necessity of latent space editing, we compare \algoname with its na\"ive nearest neighbor (NN) equivalent,
and examine three approaches.
The first approach selects each frame's nearest counterpart in the \follower motion and incorporates it into the \out motion. 
The second computes a softmax on top of the similarity scores, extracting \follower's frames according to the softmax weights. 
The third computes the similarity between the \tracked's $Q$ and the \follower's $K$ latent features, and extracts the \follower's value $V$ according to the most similar (\ie, nearest neighbor) frame.

\myparagraph{Style Transfer}
Style Transfer is a special case of motion transfer. It refers to doing a given action in a different way that represents an emotion or a physical state, such as ``happily'' or ``like a monkey''.
Typically, it retains the original action (e.g., walk), while motion transfer encompasses any change in motion motifs, including changing the action (action transfer) or modifying a subset of joints without altering the style (spatial editing).
While most motion style-transfer approaches \cite{aberman2020unpaired,jang2022motion,guo2024generative} excel with predefined style classes but struggle 
with unseen styles, our method seamlessly handles any given style motion. 
Existing style transfer methods require training and %
none of them utilizes diffusion models.
We compare with state-of-the-art MoST~\cite{kim2024most}, 
\sr{after we train it on the HumanML3D dataset.}

\myparagraph{Spatial Editing}
Spatial editing adjusts specific joints, like the arms, while preserving the overall motion.
Among spatial editing motion diffusion methods, 
MDM inpainting~\cite{tevet2023human}, as well as MEO~\cite{goel2023iterative}, use textual control and modify end features of a motion (e.g., rotation angles) without utilizing latent space. 
MDM inpainting excels at editing broad sets of joints, 
but needs refinement for individual joint edits~\cite{shafir2024human}. It relies on hard-coded joint masks, limiting edits to predefined body parts.
MEO~\cite{goel2023iterative} uses a finite hard-coded set of editing operations.
In contrast, our approach is not limited to editing a large set of joints, hard-coded masks, or a finite set of editing operations.
MDM Inpainting is the closest diffusion spatial editing work that provides code. It expects an input motion and a text, so for each pair of benchmark motions, the \tracked is used as is, and the \follower is given by its text prompt. 

\myparagraph{OOD Synthesis}
Out-of-distribution (OOD) synthesis entails generating motions that were not encountered during the training phase, posing a challenge to the network's generalization capabilities. 
For instance, when attempting to generate a dancing gorilla, the network struggles to generalize to this combination despite being trained on individual instances of dancing humans and of non-dancing gorillas. This difficulty arises from the sparse representation of such combinations in the latent space.
By applying our proposed technique, we generate latent features located sparsely away from all other features, and yet, they are denoised into a natural motion.
Existing motion diffusion methods do not transfer a given motion into an OOD one. Hence, we evaluate the OOD samples in our benchmark using style transfer and spatial editing baselines.

\myparagraph{Action Transfer}
Action transfer is the application of transferring the \outline of one action into another, say, ``running to walking'' or ``walking to crawling'', where the output would be doing the \follower's action in the same rhythm and leg order as the \tracked.
As for OOD synthesis, no current method diffuses motion into a different action. We assess action transfer samples in our benchmark using style transfer and spatial editing baselines.

\ifarxiv
    \ifarxiv
\begin{figure}
\else
\begin{figure*}
\fi
    \centering

    \ifarxiv
        \includegraphics[width=\columnwidth]{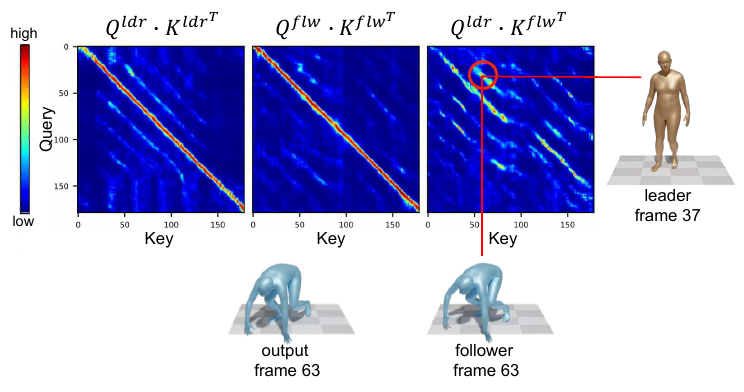}

    \else
        \includegraphics[width=.6\linewidth]    {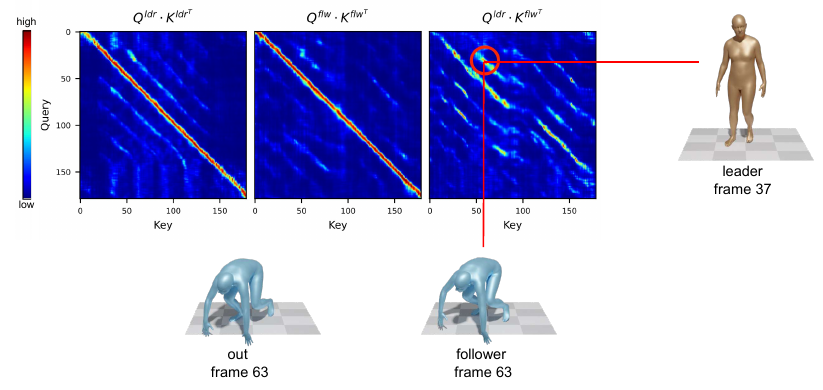}
    \fi
    \vspace{-10pt}
    \caption{\textbf{Attention maps for full motions.}
    The left and middle self-attention maps relate to a \tracked and a \follower, performing ``walking'' and ``walking like a gorilla'' motions, respectively. To the right is their \attnname map.
    For self-attention, the main diagonal indicates self-correlation and the secondary diagonals indicate the periodicity in the walks. 
    For \attnname, the high attention diagonals relate to the 
    correlation between the \tracked and the \follower.
    In the frames attributed to the circled high attention score within the \attnname map, both  \tracked and \follower depict a similar stepping pose, justifying the high activation. 
    The resulting \out frame is similar, but not identical, to the \follower's, attributed to the weighted sum that combines multiple inputs.
    All maps relate to layer $6$ and diffusion step $70$ in our backbone.
    }
    \label{fig:attention_full_motion}
    \Description[]{}  %
\ifarxiv
\end{figure}
\else
\end{figure*}
\fi
\fi

\subsection{Quantitative Results} \label{sec:qunt_res}
\begin{table}[b]
    \caption{\textbf{Comparison with baselines on the \benchname benchmark}.
    \algoname exhibits the best FID and R-precision results and comparable similarity scores. {\algoname}'s version that runs on generated motions (abbreviated ``Gen.'') has slightly better results than the one that runs on inverted ones (``Inv.''). Baselines that exhibit the highest similarity metrics scores, show poor FID and R-precision results.
    Na\"ive nearest neighbor variations attain good similarity to the follower, as they are built by features copied from it. However, they exhibit poor FID due to a jittery output.
    MoST results are too similar to the \tracked, hence fail in similarity to the \follower, and MDM Inpainting works well only when the upper body part needs to be edited, due to its fixed mask.
    \textbf{Bold} and \underline{underline} indicate best and second-best results, respectively.
    }
    \vspace{-5pt}
    \centering
    \resizebox{\columnwidth}{!}{
    \begin{tabular} {@{}l | c  c  c  c  c @{}}
        \diagbox{Model}{Metric} & FID $\downarrow$ & \makecell{R Precision \\ (top 3)} $\uparrow$ & \makecell{Leader Foot \\ Contact Sim.} $\uparrow$ & \makecell{\Follower \\ Rot. Sim.} $\uparrow$ & \makecell{\Follower \\ Loc. Sim.} $\uparrow$  \\
        \midrule
        NN motion space & 6.17 & 0.385 & \underline{0.830} & \textbf{1.00} & \textbf{1.00} \\
        \quad + softmax & 11.9 & 0.312 & 0.756 & \underline{0.994} & \underline{0.986} \\
        NN latent space  & 3.63 & 0.384 & 0.798 & 0.981 & 0.966 \\
        MoST~\shortcite{kim2024most} & 15.2 & 0.240 & 0.824 & 0.207 & 0.227 \\
        MDM Inp.~\shortcite{tevet2023human}  & 3.51 & 0.213 & \textbf{1.00} & 0.244 & 0.329 \\
        \algoname Gen. (Ours)  & \textbf{2.33} & \underline{0.439} & 0.816 & 0.993 & 0.972 \\  %
        \algoname Inv. (Ours)  & \underline{2.50} & \textbf{0.490} & 0.793 & 0.933 & 0.856 \\  %
        \bottomrule
    \end{tabular}
    } %
    \label{tab:hml_quant_res}
\end{table}

In \cref{tab:hml_quant_res} we compare \algoname with the baselines mentioned above, 
using the \benchname benchmark.
Our work excels in FID and R-precision while achieving comparable similarity scores. However, baselines with the highest similarity scores show inferior FID and R-precision results. \algoname's version operating on generated motions outperforms the one on inverted motions.

\subsection{Qualitative Results} \label{sec:qual_res}
\ifarxiv
    \begin{figure*}
    \centering
    \includegraphics[width=\linewidth]{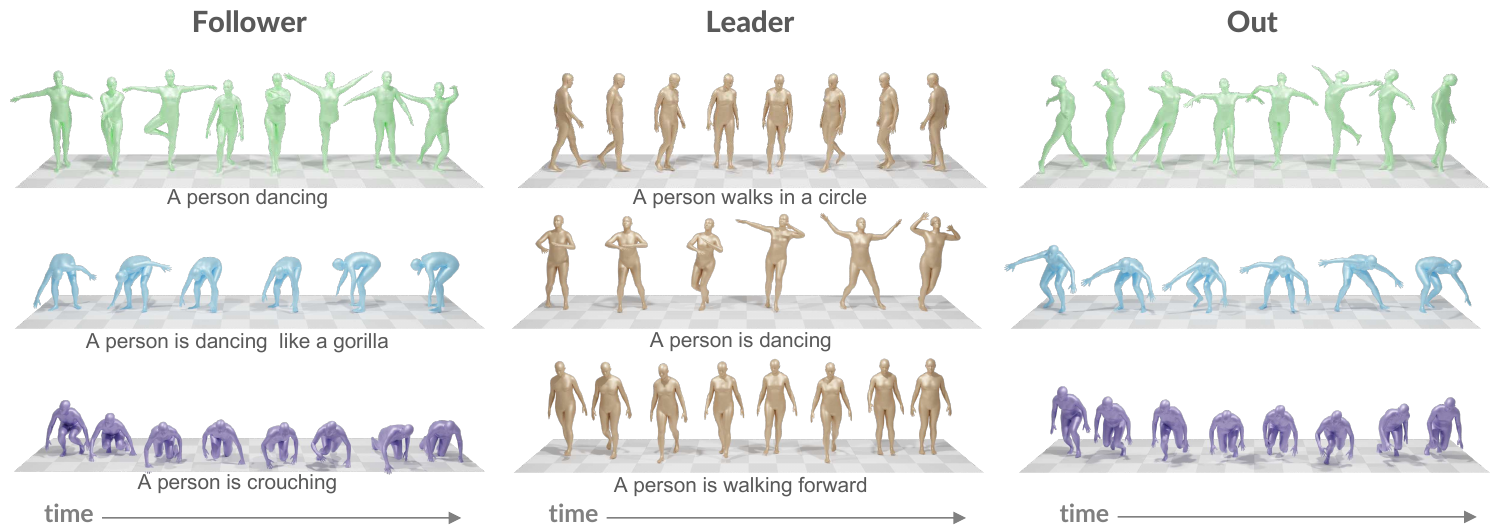}
    \vspace{-10pt}
    \caption{\textbf{Qualitative Results.} The results here demonstrate the transfer of the \tracked motion's \outline, such as the sequence and pace of steps, to the \follower while preserving the subtle nuances of the \follower's motion.
    The first and third rows relate to the special case of an action transfer. In the second row, we illustrate an out-of-distribution synthesis scenario; initially, the \follower lacks a dancing motion, despite the text prompt. However, after the transfer process, the synthesized motion imitates a gorilla dancing, as specified.}
    \vspace{-5pt}
    \label{fig:qual_res}
    \Description[]{}  %
\end{figure*}
\fi
Our supplementary video reflects the quality of our results. 
It presents multiple transferred motions and comparisons to other baselines.
It shows that the first nearest neighbor approach results in jittery outcomes, the second produces unnatural outcomes with foot sliding, MDM Inpainting cannot generalize beyond its spatial editing expertise, and MoST struggles to generalize on unseen styles.
\algoname offers a versatile motion transfer technique expressed in  ~\cref{fig:teaser,fig:qual_res} where we see how it facilitates tasks unified by the core concept of transferring a motion \outline from a \tracked to a \follower while retaining the \follower's unique \motifs. Arranged from top to bottom, \cref{fig:teaser} demonstrates special cases of spatial editing, style transfer, and action transfer. \cref{fig:qual_res} illustrates action transfer, out-of-distribution synthesis, and another action transfer.

\subsection{Ablation}

\begin{table}[b]
    \caption{\textbf{Layers and steps ablation.}
    The table displays representative results of the variations we experimented with diffusion steps and layers in which the self-attention features are manipulated.  
    The top row displays the configuration of our selected model, where we apply \attnname for self-attention layers 2-12 and for diffusion steps 90 to 10 (out of 100). In the middle block, we maintain our best layer configuration and test various diffusion step ranges. In the bottom block, we maintain our best diffusion steps configuration and experiment with the range of layers.
    To select the best configuration, we prioritized FID and R-Precision over the other metrics.
    }
    \vspace{-5pt}
    \centering
    \resizebox{\columnwidth}{!}{
    \begin{tabular} {@{} l l | c  c  c  c c @{}}
         Layers   & \makecell{Diffusion \\ Steps} & FID $\downarrow$ & \makecell{R Precision \\ (top 3)} $\uparrow$ & \makecell{Leader Foot \\ Contact Sim.} $\uparrow$ & \makecell{\Follower \\ Rot. Sim.} $\uparrow$ & \makecell{\Follower \\ Loc. Sim.} $\uparrow$  \\
        \midrule
2 - 12 & 10 - 90 & \textbf{2.334} & \textbf{0.439} & 0.816 & \textbf{0.993} & 0.972 \\ %
\midrule
2 - 12 & 20 - 80 & 3.028 & 0.412 & 0.820 & \textbf{0.993} & 0.973 \\ %
2 - 12 & 15 - 90 & 2.833 & 0.406 & 0.817 & 0.992 & \textbf{0.975} \\ %
2 - 12 & 20 - 70 & 2.867 & 0.416 & 0.821 & 0.991 & 0.973 \\ %
\midrule
4 - 9 & 10 - 90 & 4.063 & 0.373 & 0.795 & 0.978 & 0.971 \\ %
5 - 11 & 10 - 90 & 2.971 & 0.393 & \textbf{0.837} & 0.989 & 0.967 \\ %
4 - 10 & 10 - 90 & 3.098 & 0.404 & 0.821 & 0.991 & 0.974 \\ %
        \bottomrule
    \end{tabular}
    } %
    \label{tab:hml_ablation}
\end{table}

\begin{table}[b]
    \caption{\textbf{Text prompt ablation.} In each row, a different textual prompt is utilized for the \out motion. Results indicate that using the same text as the \follower yields the best outcomes. 
    }
    \vspace{-5pt}
    \centering
    \resizebox{\columnwidth}{!}{
    \begin{tabular} {@{} l | c  c  c  c c @{}}
         \diagbox{Prompt}{Metric} & FID $\downarrow$ & \makecell{R Precision \\ (top 3)} $\uparrow$ & \makecell{Leader Foot \\ Contact Sim.} $\uparrow$ & \makecell{\Follower \\ Rot. Sim.} $\uparrow$ & \makecell{\Follower \\ Loc. Sim.} $\uparrow$  \\
        \midrule

Same as \follower & \textbf{2.572} & \textbf{0.434} & 0.814 & \textbf{0.994} & \textbf{0.975} \\ %
None & 3.182 & 0.410 & 0.817 & 0.986 & 0.948 \\ %
``A person'' & 3.282 & 0.391 & \textbf{0.824} & 0.986 & 0.947 \\ %

        \bottomrule
    \end{tabular}
    } %
    \label{tab:hml_prompt_ablation}
\end{table}

Our initial study focuses on identifying the appropriate layers and diffusion steps for which \attnname should be applied. We denote the range of layers and diffusion steps where \attnname is applied as $[s\mbox{-}layer,e\mbox{-}layer]$ and $[s\mbox{-}step,e\mbox{-}step]$, respectively.
We have tested approximately 200 configurations, with varying values of $s\mbox{-}layer$, $e\mbox{-}layer$, $s\mbox{-}step$, and $e\mbox{-}step$.
\Cref{tab:hml_ablation} displays representative results of the variations we experimented with. 

Another ablation test examines the textual prompt used during the synthesis of the \out motion. Results in \cref{tab:hml_prompt_ablation} confirm that our choice to replicate the \follower's prompt is optimal.

\section{Conclusion, Limitations and Future Work} \label{sec:conclusion}

We have explored and leveraged the powerful mechanism of self-attention, within pre-trained motion diffusion models.
Our study has resulted in a novel approach for motion editing in diffusion models, which we have demonstrated through motion transfer.

Our zero-shot and unpaired methodology, named \algoname (\algonamelong), demonstrates promising capabilities in transferring the \outline of a \tracked motion onto a \follower one while preserving the subtle motifs of the \follower. 
This enables a variety of sub-tasks such as out-of-distribution motion synthesis, style transfer, action transfer, 
and spatial editing. 
By harnessing motion inversion, \algoname extends its applicability to both real and generated motions.

Experimental results demonstrate the merits of our approach compared to the baselines.
Notably, it excels in applying functionalities at inference time without additional training.
Our findings shed light on the importance of attention in human motion and lay the groundwork for future advancements in the field.

The primary limitation of our work is the difficulty in transferring a motion when basic \outline elements of the \tracked motion are lacking in the \follower.
For example, if the \tracked motion depicts walking while the \follower remains stationary, then the \out motion will also be stationary, receiving no input from the \tracked.
However, this limitation may be inherent to the task description itself, as our objective is primarily to transfer motions that share some degree of commonality in their {\outline}s. %
Another current limitation is that once a \tracked and a \follower motions are determined, there is no diversity in the \out motions obtained from them. 
This limitation arises from the deterministic nature of the DDIM~\cite{song2020denoising} diffusion model we are using. In the future, output diversity could be achieved using non-deterministic models (e.g., DDPM).

A possible future direction we offer here is to explore latent feature layers other than self-attention, \eg, cross-attention and feed-forward. 
Additionally, our findings on motion \motifs preservation could be applied to personalization applications, a domain popular in imaging~\cite{gal2023image} but currently lacking in the motion domain.

\ifanonymous\else
    \vspace{-5pt}

\section{Acknowledgments}
We thank 
Rinon Gal and Or Patashnik for fruitful discussions.
This research was supported in part by the Israel Science Foundation (grants no. 2492/20 and 3441/21), Len Blavatnik and the Blavatnik family foundation, and the Tel Aviv University Innovation Laboratories (TILabs).

\fi

\bibliographystyle{ACM-Reference-Format}
\bibliography{92_bibiliography}

\ifappendix
    \section*{\textbf{Appendix}}
    \appendix
    This \ifappendix{Appendix }\else{Supplementary }\fi  adds details on top of the ones given in the main paper. While the main paper stands on its own, the details given here may shed more light. 

In \cref{sec:prelim_supp} we provide more details regarding the preliminaries of our work: motion representation, and the models MDM, DDPM, and DDIM.
\Cref{sec:exp_supp} elaborates on our experiments, in particular the benchmark, metrics, hypermparameters, and additional implementation details. 

\section{Preliminaries - More Details} \label{sec:prelim_supp}

\myparagraph{Motion Representation}
Recall that $N$ denotes the number of frames in a motion sequence, $F$ denotes the length of the features describing a single frame, $J$ denotes the number of skeletal joints, and $X\in \R^{N\times F}$ denotes a motion.
Each feature is redundantly represented with the joint angles, positions, velocities, and foot contact~\cite{guo2022generating}.
Each single pose is defined by 
\[
(\dot{r}^{a}, \dot{r}^x, \dot{r}^z, r^y, j^p, j^r, j^v, c^f) \in \R^F,
\]
where $\dot{r}^{a} \in \R$ is the root angular velocity along the Y-axis. $\dot{r}^x, \dot{r}^z \in \R$ are root linear velocities on the XZ-plane, and $r^y \in \R$ is the root height. $j^p \in \R^{3(J-1)}$, $j^r \in \R^{6(J-1)}$ and $j^v \in \R^{3J}$ are the local joint positions, velocities, and rotations relative to the root, and $c^f \in \R^4$ are binary features denoting the foot contact labels for four foot joints (two for each leg).

\myparagraph{MDM, DDPM and DDIM}

Recall that MDM~\cite{tevet2023human} does human motion synthesis and editing and that it uses DDPMs~\cite{ho2020denoising}. In the following, we briefly describe the mechanism of DDPM.

An input motion $x_0$, is subjected to a Markov noise process consisting of T steps, resulting in the sequence $\{x_t\}_{t=0}^T$,  such that 
\begin{equation}
q(x_{t} | x_{t-1}) = \nn(\sqrt{\alpha_{t}}x_{t-1},(1-\alpha_{t})I),
\end{equation}
where $\alpha_{t} \in (0,1)$ are constant hyper-parameters. When  $\alpha_{t}$ is small enough, we can approximate $x_{T} \sim \nn(0,I)$.

$x_0$ can be modeled via the reversed diffusion process by gradually cleaning $x_{T}$, using a generative network $p_\theta$. MDM~\cite{tevet2023human} predicts the input motion, denoted $\hat{x}_{0}$, rather than $\epsilon_t$, such that   
$\hat{x}_0 = p_\theta(x_t, t)$. Then, the widespread diffusion loss is applied:

\begin{equation}
\Loss_\text{simple} = \E_{t \sim [1,T]}\norm{ x_0 - p_\theta(x_t, t)}_2^2.
\end{equation}

During inference, synthesis iterates from pure noise $x_T$. 
In each iteration, the denoising network $p_\theta$ predicts a clean version of the current sample $x_t$. The predicted clean sample $\hat{x}_{0}$ is then ``re-noised" to create the next sample $x_{t-1}$, repeatedly until  $t=0$.

Denoising Diffusion Implicit Models (DDIM)~\cite{song2021denoising} enable a non Markovian, deterministic, version of DDPMs.
DDIM can be applied on a network pre-trained with DDPM, for either accelerated inference, or inversion.
In this work, we employ DDIM for inversion and deterministic inference, to reconstruct inverted motions precisely to their original form.

\section{Experiments - More details} \label{sec:exp_supp}
\subsection{Benchmark}
The complete list of filter terms used to create the \benchname benchmark is as follows.
For each pair in the benchmark, the \tracked contains one of the following locomotion key words: ``run'', ``walk'', ``jump'', ``danc''. The \follower contains one of the aforementioned locomotion key words, plus one style or action key word, from the list ``gorilla'', ``drunk'', ``robot'', ``chicken'', ``frog'', ``monkey'', ``style'', ``like'', ``old'', ``child'', 
``raise'', ``clap'', ``wav'', ``kick'', ``punch'', ``push'', ``pull''. 

Following are several examples of pairs from the \benchname benchmark. Each example is identified by its text prompt and its index number in the HumanML3D dataset.
\begin{itemize}
    \item \Tracked: ``a person walks towards the left making a wide 's' shape'' (\#009488). \Follower: ``a person walks in a clockwise circle and raises their hand to their face to yawn'' (\#004222). 
    \item \Tracked: ``a person runs and then jumps'' (\#007291). \Follower: ``the drunk guy struggles to walk down the street'' (\#005037). 
    \item \Tracked: ``it is a person walking backwards'' (\#007199). \Follower: ``person is walking with his arms out like he is balancing'' (\#010823). 
\end{itemize}

\subsection{Metrics}

\myparagraph{FID~\cite{guo2022generating,heusel2017gans}}
We use this metric to assess the quality of the \out motions. 
The Fréchet Inception Distance (FID) evaluates the similarity between the distribution of real motions and synthesized ones 
We consider the benchmark dataset as the ground truth distribution and extract features from both real motions in the benchmark and generated \out motions. 
Motions are deemed of high quality if they exist within the manifold of the ground truth.
 
\myparagraph{R-precision~\cite{guo2022generating}}

We use this metric to assess the alignment between each \out motion and the motion motifs of their associated \follower motions. 
R-precision uses a latent space shared between text and motion to measure the distance between motion and text embeddings. 
Hence, it measures whether a motion reflects some text prompt.
The text prompt of each \out motion is copied from the \follower used when generating it, so a successful match means the \out motion adheres to the motion pattern of the \follower.

For each generated motion, we create a description pool consisting of its ground-truth text description and 31 randomly chosen, unrelated descriptions from the test set. 
We then calculate and rank the Euclidean distances between the motion feature and the text feature of each description in the pool. We measure success by the average accuracy at the top-1, top-2, and top-3 positions. A successful retrieval occurs when the ground truth entry ranks in the top-k candidates.

\myparagraph{Foot Contact Similarity}
This metric assesses how well the \out motion retains the locomotion rhythm of the \tracked motion. 
Successful motion transfer should preserve the rhythm from \tracked to \out motion, so the foot contact labels of the two motions should closely align. 
We measure this similarity score by comparing the foot contact data within the HumanML3D motion features. 
Matching foot contact labels are counted as hits while differing labels are counted as misses. 
The metric score is determined by the rate of hits.

\myparagraph{Similarity to \Follower Rotations}
This metric evaluates the fidelity of the \out motion to the subtle motifs found in the \follower. Specifically, it assesses the resemblance of rotations within the \out motion to either the \tracked or \follower motions. Given that subtle nuances are typically expressed in most joints' rotations, we anticipate observing a higher degree of similarity to the \follower motion.

To compute this metric, the initial step is to identify, for each frame, the nearest neighbor in both the \tracked and \follower motions. This entails identifying the closest match for each frame.

Once the nearest neighbors are identified, the metric calculates the rate of frames where the similarity to the \follower's nearest neighbor surpasses that of the \tracked's. Essentially, it evaluates the proportion of rotation frames where the outbound motion exhibits greater alignment with the \follower's motion compared to the tracked one.

\myparagraph{Similarity to \Follower Locations}
This metric also evaluates the fidelity of the \out motion to \follower's motifs.
Similar to the previous metric, it calculates the nearest neighbors for each frame, but this time for the locations relative to the root. As with the previous metric, we expect a greater degree of similarity to the \follower motion.

\subsection{Implementation Details}

\myparagraph{Backbone and hyperparameters}
\begin{table}[htp]
    \caption{\textbf{Hyperparameters used for our backbone.} Our backbone is a variation of MDM, with the hyperparameters listed here.}
    \vspace{-10pt}
    \centering
    \begin{tabular}{ l | c }
        Name &  Value \\
        \midrule
        
         \underline{Model} & \\
             \quad Architecture & Transformer Decoder \\ %
             \quad layers & 12  \\ %
             \quad latent dim & 512 \\ %

         \underline{Diffusion} & \\
             \quad diffusion steps & 100 \\ %
             \quad noise schedule & cosine \\ %
             \quad guidance scale & 2.5 \\ %

         \underline{Training} & \\
             \quad batch size & 32 \\  %
             \quad lr & 0.0001 \\  %
             \quad dropout & 0.1 \\  %
             \quad num steps & 600000 \\  %
             \quad warmup steps & 0 \\
             \quad weight decay & 0 \\
             \quad seed & 10 \\  %
        
        \bottomrule
    \end{tabular}
    \label{tab:backbone_hyperparams}
    \ifarxiv
    \else
    \vspace{-5pt}
    \fi
\end{table}

\begin{table}[htp]
    \caption{\textbf{Hyperparameters used during inference when applying \algoname.}}
    \vspace{-10pt}
    \centering
    \begin{tabular}{ l | c }
        Name &  Value \\
        \midrule

         \underline{Applying \attnname} & \\
             \quad Layers & range(1, 12) \\ %
             \quad Diffusion steps & range(10, 91) \\ %
         \underline{Text prompt} & \\
             \quad \Out motion & Same as \follower \\ %
      
        \bottomrule
    \end{tabular}
    \label{tab:mot_trans_hyperparams}
    \ifarxiv
    \else
    \vspace{-5pt}
    \fi
\end{table}

The backbone we use for reporting experimental results is the transformer decoder variation of MDM~\cite{tevet2023human}. We retrain it on the HumanML3D~\cite{guo2022generating} dataset, with the hyperparameters shown in ~\cref{tab:backbone_hyperparams}. In particular, we slightly change the architecture such that the text embedding is given twice: once as an extra-temporal token like in the transformer encoder variation, and once as separate word tokens using cross-attention. 

Additionally, ~\cref{tab:mot_trans_hyperparams} provides hyperparameters used during inference when applying \algoname.

\myparagraph{Controlling directions during motion transfer}

A natural challenge arises when the locomotion direction of the \tracked motion differs from that of the \follower's.
In such cases, the \out motion would retain the \outline of the \tracked, such as the timing of steps, but would proceed in a different direction.
Fortunately, the solution is straightforward. 
For generated motions, we create multiple followers, each generated with a different seed, for the given text prompt.
\algoname is then applied to the concatenation of all {\follower}s together.
Our experiments show that generating just a few followers ensures that the output motion closely follows the direction of the follower.
For inverted motions, using different seeds would not help as the inversion by DDIM is deterministic. 
Our solution is to rotate the given motion around the vertical axis in several rotation angles. 
Similar to the generative case, we apply our framework to the concatenation of all rotated followers.
Consequently, each \tracked's frame attends to the frames of all the {\follower}s and achieves the highest attention scores with those that have directions similar to its own. 
This facilitates near-accurate direction transfer.
In practice, we have found that this solution is nearly equivalent to copying the root rotation value from the \tracked to the \out motion.
For evaluation, we used the latter, to accelerate computation time.
\fi
\end{document}